\documentclass[10pt,twocolumn,letterpaper]{article}
\usepackage[pagenumbers]{cvpr} 

\usepackage{capt-of}
\usepackage{xspace}
\usepackage{xcolor}
\usepackage{enumitem}
\usepackage{amsmath,amsthm,amssymb,amsfonts,dsfont,pifont,bm,bbm,mathrsfs,mathtools,nicefrac,extarrows,relsize}
\usepackage{algorithm,algpseudocode,listings}
\usepackage{booktabs,multirow,adjustbox,diagbox,threeparttable,tabularray,setspace}

\definecolor{cvprblue}{rgb}{0.21,0.49,0.74}
\usepackage[pagebackref,breaklinks,colorlinks,citecolor=cvprblue,bookmarks=false]{hyperref}
\usepackage{wrapfig}
\usepackage[capitalize]{cleveref}

\crefname{section}{Sec.}{Secs.}
\Crefname{section}{Section}{Sections}
\crefname{appendix}{App.}{Apps.}
\Crefname{appendix}{Appendix}{Appendices}
\crefname{table}{Tab.}{Tabs.}
\Crefname{table}{Table}{Tables}
\crefname{figure}{Fig.}{Figs.}
\Crefname{figure}{Figure}{Figures}
\crefname{equation}{Eq.}{Eqs.}
\Crefname{equation}{Equation}{Equations}
\crefname{theorem}{Thm.}{Thms.}
\Crefname{theorem}{Theorem}{Theorems}
\crefname{lemma}{Lem.}{Lems.}
\Crefname{lemma}{Lemma}{Lemmas}
\crefname{remark}{Rem.}{Rems.}
\Crefname{remark}{Remark}{Remarks}
\crefname{corollary}{Cor.}{Cors.}
\Crefname{corollary}{Corollary}{Corollaries}
\crefname{algorithm}{Alg.}{Algs.}
\Crefname{algorithm}{Algorithm}{Algorithms}
\definecolor{cellred}{RGB}{213, 123, 101}
\definecolor{cellgreen}{RGB}{0, 205, 0}
\definecolor{cellblue}{RGB}{54, 125, 189}
\definecolor{codegreen}{rgb}{0,0.6,0}
\definecolor{codegray}{rgb}{0.5,0.5,0.5}
\definecolor{codepurple}{rgb}{0.58,0,0.82}
\definecolor{backcolour}{rgb}{1.0,1.0,1.0}
\lstdefinestyle{mystyle}{
    backgroundcolor=\color{backcolour},
    commentstyle=\color{codegreen},
    keywordstyle=\color{magenta},
    numberstyle=\tiny\color{codegray},
    stringstyle=\color{codepurple},
    basicstyle=\ttfamily\scriptsize,
    breakatwhitespace=false,
    breaklines=true,
    captionpos=b,
    keepspaces=true,
    numbers=left,
    numbersep=5pt,
    showspaces=false,
    showstringspaces=false,
    showtabs=false,
    tabsize=2
}
\newcommand{\methodname}{PlanarSplatting}
\newcommand{\method}{\textit{\methodname}\xspace}

\title{PlanarSplatting: Accurate Planar Surface Reconstruction in 3 Minutes}

\author{Bin Tan\textsuperscript{1} \quad Rui Yu\textsuperscript{2} \quad Yujun Shen\textsuperscript{1} \quad Nan Xue\textsuperscript{$\dagger$,1} \\ \\
\textsuperscript{1}Ant Group  \quad \textsuperscript{2}University of Louisville}

\begin{document}

\twocolumn[{
\renewcommand\twocolumn[1][]{#1}
\maketitle
\begin{center}
    \includegraphics[width=0.99\linewidth]{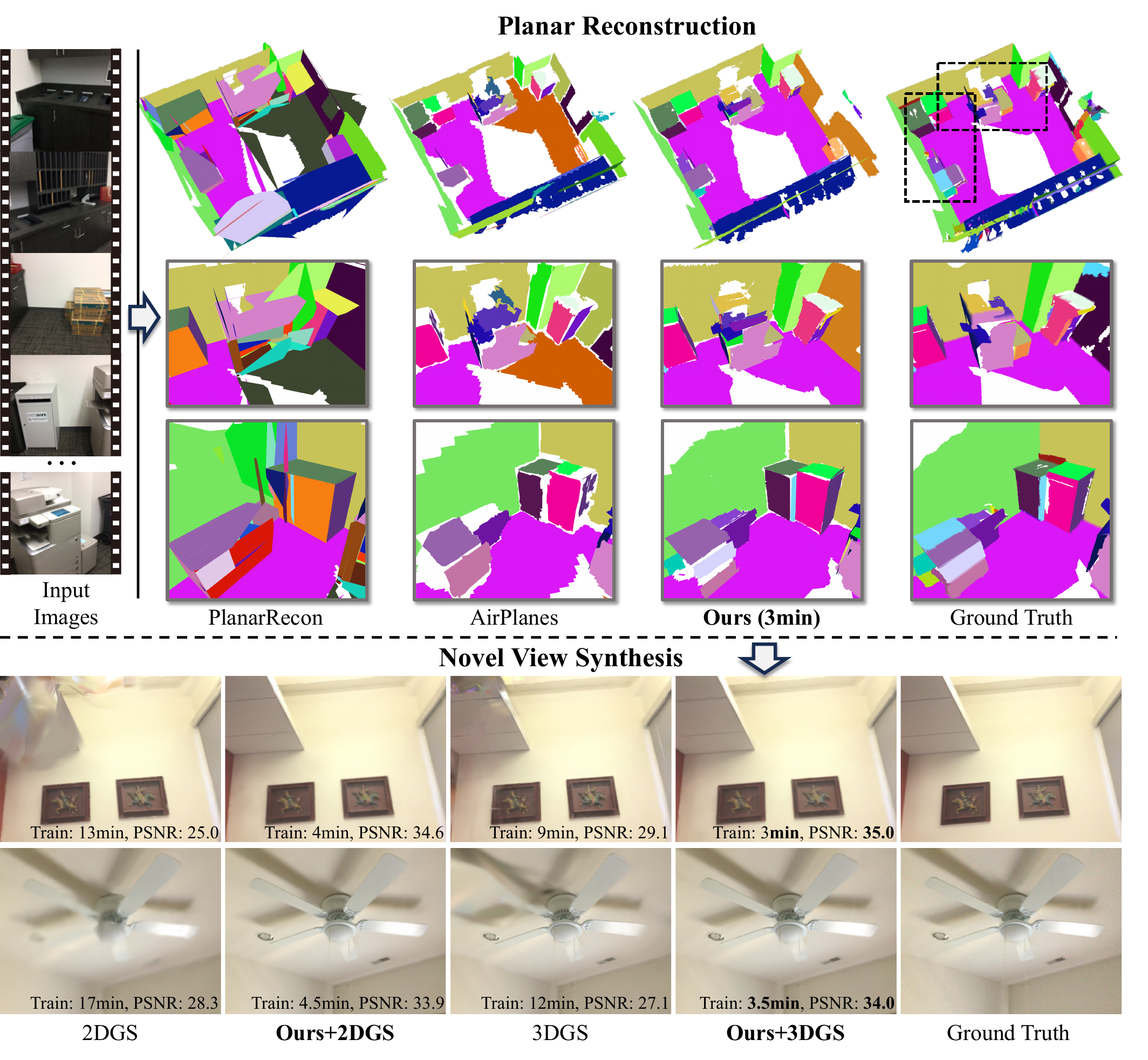}
    \captionsetup{type=figure}
    \vspace{-0.5cm}
    \caption{%
        We introduce \method, a fast and accurate optimization-based planar reconstruction method for indoor scenes. \textbf{Top (Planar Reconstruction)}: 
        We show our planar reconstruction results on the ScanNetV2~\cite{scannet-DaiCSHFN17} dataset achieved in 3 minutes. 
        Compared to prior art PlanarRecon~\cite{PlanarRecon-XieGYZJ22} and AirPlanes~\cite{AirPlanes-WatsonASQABFV24}, our \method reconstructs more complete and detailed 3D planes. \textbf{Bottom (Novel View Synthesis)}: We show that our \method can be seamlessly integrated with recent Gaussian Splatting methods (e.g., 3DGS~\cite{ThreeDGS-KerblKLD23} and 2DGS~\cite{TwoDGS-HuangYC0G24}) to achieve improved rendering results in indoor scenes while requiring significantly less optimization time.
    }
    \label{fig:teaser}
\end{center}
}]

\begin{abstract}
This paper presents \method, an ultra-fast and accurate surface reconstruction approach for multiview indoor images. We take the 3D planes as the main objective due to their compactness and structural expressiveness in indoor scenes, and develop an explicit optimization framework that learns to fit the expected surface of indoor scenes by splatting the 3D planes into 2.5D depth and normal maps. 
As our \method operates directly on the 3D plane primitives, it eliminates the dependencies on 2D/3D plane detection and plane matching and tracking for planar surface reconstruction. Furthermore, the essential merits of plane-based representation plus CUDA-based implementation of planar splatting functions, \method reconstructs an indoor scene {\bf in 3 minutes} while having significantly better geometric accuracy. Thanks to our ultra-fast reconstruction speed, the largest quantitative evaluation on the ScanNet and ScanNet++ datasets over hundreds of scenes clearly demonstrated the advantages of our method. We believe that our accurate and ultrafast planar surface reconstruction method will be applied in the structured data curation for surface reconstruction in the future. The code of our CUDA implementation will be publicly available. Project page can be found \href{https://icetttb.github.io/PlanarSplatting/}{here}.

\end{abstract}

\section{Introduction}\label{sec:intro}
We humans are long-term immersed in structured indoor scenes, ranging from bedrooms to offices. This fact has ignited a wealth of studies for reconstructing the indoor environment with various 3D structures such as lines~\cite{Line3D++,limap,NEAT-TXDX0S24}, planes~\cite{planercnn-0012KGFK19,planeAE-YuZLZG19,PlanarRecon-XieGYZJ22,NOPE-SAC-TanXWX23,Sparseplanes-Jin0OF21,GeometricReasoning-LeeHK09} and blocks~\cite{BlocksWorldRevisited,DBW-abs-2307-05473}. Among them, the 3D plane is the most common representation because of its simplicity and completeness in describing physical surfaces, thus motivating us to study the problem of 3D reconstruction for indoor scenes using 3D planes, \ie, planar 3D reconstruction.

Planar 3D reconstruction has been extensively studied for years as a model fitting problem, in which a 3D scene geometry (\eg, point clouds, or meshes) was assumed to be known and the main goal is fitting the scene in a set of 3D planes~\cite{PointFit-SommerSGCB20, fastplane-PoppingaVBP08,XuWWY21}. 
Recently, the paradigm has been gradually simplified in image-based solutions in single-view and multiview 3D reconstruction, eliminating the acquisition of known 3D scene geometry. 
To make the problem trackable, existing methods were extensively based on the image-level characterization of 2D/3D planes. That is to say, those methods have to detect 3D planes for each input image, match or track planes in across viewpoints, and finally reconstruct and merge 3D planes as the compact 3D representation of indoor scenes. 
Recent PlanarRecon~\cite{PlanarRecon-XieGYZJ22} attempted to address these problems end-to-end by learning a 3D volume from monocular videos. Then, 3D planes can be detected, tracked, and fused in a consistent 3D space. 

In fact, image-based planar 3D reconstruction mostly followed keypoint-based 3D reconstruction pipelines and treated planes as a special kind of visual features. However, because of the essential difference between the local point features and regional plane masks in image space, we argue, the existing approaches did not fully leverage the advantages of planar representations. Some evidences could be observed from the results of PlanarRecon~\cite{PlanarRecon-XieGYZJ22}, which are usually very coarse and lose many details of the scene. 

In this paper, we are going to address the issues remained in image-based planar 3D reconstruction, aiming at obtaining a complete, structural, and compact indoor scene reconstruction from multi-view images. 
Our main idea is approximating the indoor scenes with a collection of solid 3D planar primitives from multi-view input images, directly optimizing them to have consistent 3D planes, but eliminating any suboptimal precomputing of plane primitives (\eg, plane masks).
We introduce~\method that explicitly optimizes rectangular plane primitives in 3D space by differentiably splatting them into 2.5D depth and normal maps. Thanks to our well-designed plane splatting function, \method effectively leverage monocular geometry cues from modern foundational models~\cite{Metric3D-CYWCS23,Metric3Dv2,omnidata-EftekharSMZ21} for accurate plane optimization. 
As shown at the top of Figure~\ref{fig:teaser}, the final high-quality planar surface can be reconstructed by simply merging similar 3D plane primitives without any plane annotations for supervision or matching/tracking operations.

As our \method is directly designed on the 3D plane primitives and efficiently implemented with CUDA, it can be seamlessly integrated with recent Gaussian Splatting (GS) methods for high-quality indoor novel view synthesis (NVS). As shown at the bottom of~\cref{fig:teaser}, benefiting from our fast and accurate scene reconstruction (within 3 minutes), GS-based methods can be well-initialized and optimized without densification, leading to better rendering results and significantly less training time. 
It demonstrates the strong potential of our \method to promote the unity of reconstruction and novel view synthesis for the representation of indoor scenes.

In the experiments, our \method shows its powerful ability for accurate indoor planar surface reconstruction on two real-world indoor datasets including ScanNetV2~\cite{scannet-DaiCSHFN17} and ScanNet++\cite{scannetpp-YeshwanthLND23} on hundreds of scenes. Furthermore, we show that with the combination of our \method and GS-based methods (\eg, 3DGS~\cite{ThreeDGS-KerblKLD23} and 2DGS~\cite{TwoDGS-HuangYC0G24}), we can effectively improve the rendering quality with less training time and fewer points.

\section{Related Work}\label{sec:related}
\noindent\textbf{Indoor Planar Reconstruction.}
Reconstructing indoor scenes with 3D plane primitives has been studied for a long time~\cite{FurukawaCSS09-1,FurukawaCSS09-2,XiaoF14,GallupFP10,SinhaSS09}. Traditionally, it is usually achieved by deducing and fitting the plane geometry directly from 3D data (\eg, point clouds~\cite{fastplane-PoppingaVBP08,OrientedPlane-SunM19,RAPter-Mitra15,PointFit-SommerSGCB20} and line clouds~\cite{SurfLine-LangloisBM19}), or from 2D single-view images with strict scene assumptions (\eg, the Manhattan World constrain)~\cite{Towardsdetection-MicusikWV08,GeometricReasoning-LeeHK09,LS3D-QianRE18} which seriously limited their application. In recent years, some learnable-based methods are proposed to formulate this problem as single-view 3D plane segmentation~\cite{PlaneNet-LiuYCYF18,planercnn-0012KGFK19,planeAE-YuZLZG19,PlaneTR-Tan0B0X21,PlaneRecTR-ShiZ023, PlaneAC-ZhangYFM24} and cross-view plane instance matching~\cite{Sparseplanes-Jin0OF21,PlaneFormers-AgarwalaJRF22,NOPE-SAC-TanXWX23}. Although impressive results have been achieved, these works are hard to extend to multiple views. PlanarRecon~\cite{PlanarRecon-XieGYZJ22} was the first end-to-end work proposed to deal with holistic indoor plane recovery by learning a plane-related 3D volume from posed RGB videos. Then, 3D planes can be extracted, tracked, and fused from the learned 3D volume incrementally. Most recently, AirPlanes~\cite{AirPlanes-WatsonASQABFV24} developed a two-step method that first reconstructed a dense scene mesh and then learned consistent 3D plane embeddings from 2D plane embeddings for plane extraction from the dense mesh. 
Note that the aforementioned learning-based methods require 2D/3D plane annotations as their supervision, leading to performance bottlenecks due to the difficulty of obtaining a large scale of plane annotations.
In contrast, we proposed \method to reconstruct accurate and complete indoor planar surface by directly optimizing a set of solid 3D plane primitives from posed multi-view images without any extra plane detection or matching operations. Benefiting from our differentiable planar primitive rendering, \method can directly leverage monocular depth/normal cues from modern foundation models for optimization without plane annotations.

\noindent\textbf{Primitive-based Scene Representation.}
Optimizing explicit primitives such as points~\cite{zhang2023papr,ThreeDGS-KerblKLD23,TwoDGS-HuangYC0G24,DSS-WangSWOS19}, volumes~\cite{MVP-LombardiSSZSS21}, and superquadric~\cite{DBW-abs-2307-05473} to represent the 3D scene has been studied for a long time. The core of these methods is to design a differentiable rendering process to optimize the attributes of primitives by gradient descent. Among them, a typical paradigm is to render images from primitives with splatting techniques which is realized with radial basis functions defined on the primitive (\eg, the Gaussian function). Inspired by these works, we proposed \method to optimize solid 3D plane primitives in a differentiable rendering manner with a novel shape-aware plane splatting function to better fit the scene geometry. Benefiting from our efficient CUDA implementation, \method can reconstruct the accurate indoor planar scene within 3 minutes and empower recent Gaussian-based works (\eg, 3DGS~\cite{ThreeDGS-KerblKLD23}) to further improve the rendering quality of novel views as shown in~\cref{fig:teaser}.

\begin{figure}
  \centering
  \includegraphics[width=0.5\linewidth]{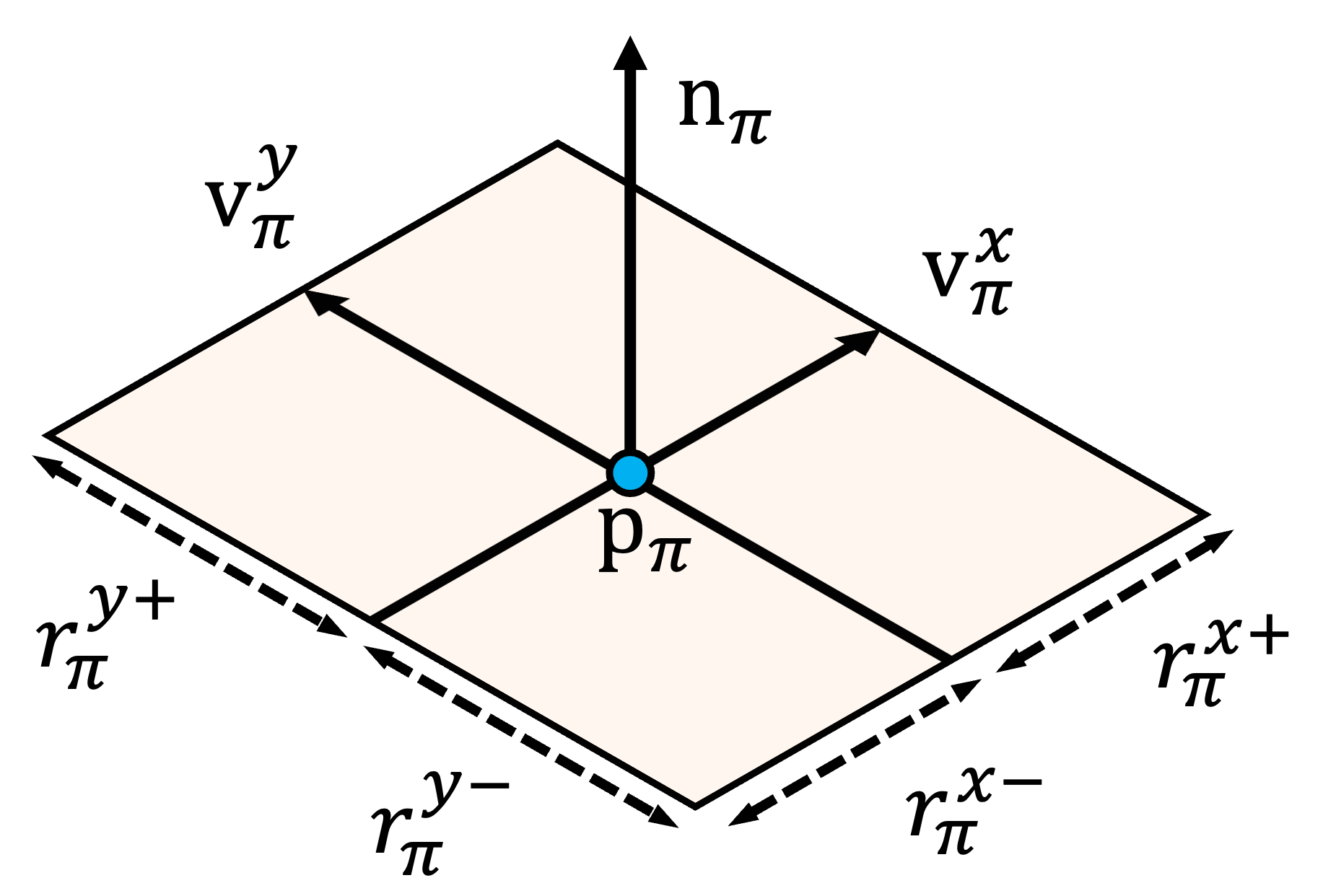}
  \caption{Representation of our 3D plane primitive with learnable shape parameters including plane center, plane radii, and plane rotation.}
  \label{fig:plane_rep}
\end{figure}

\begin{figure*}[!t]
  \centering
   \includegraphics[width=0.95\linewidth]{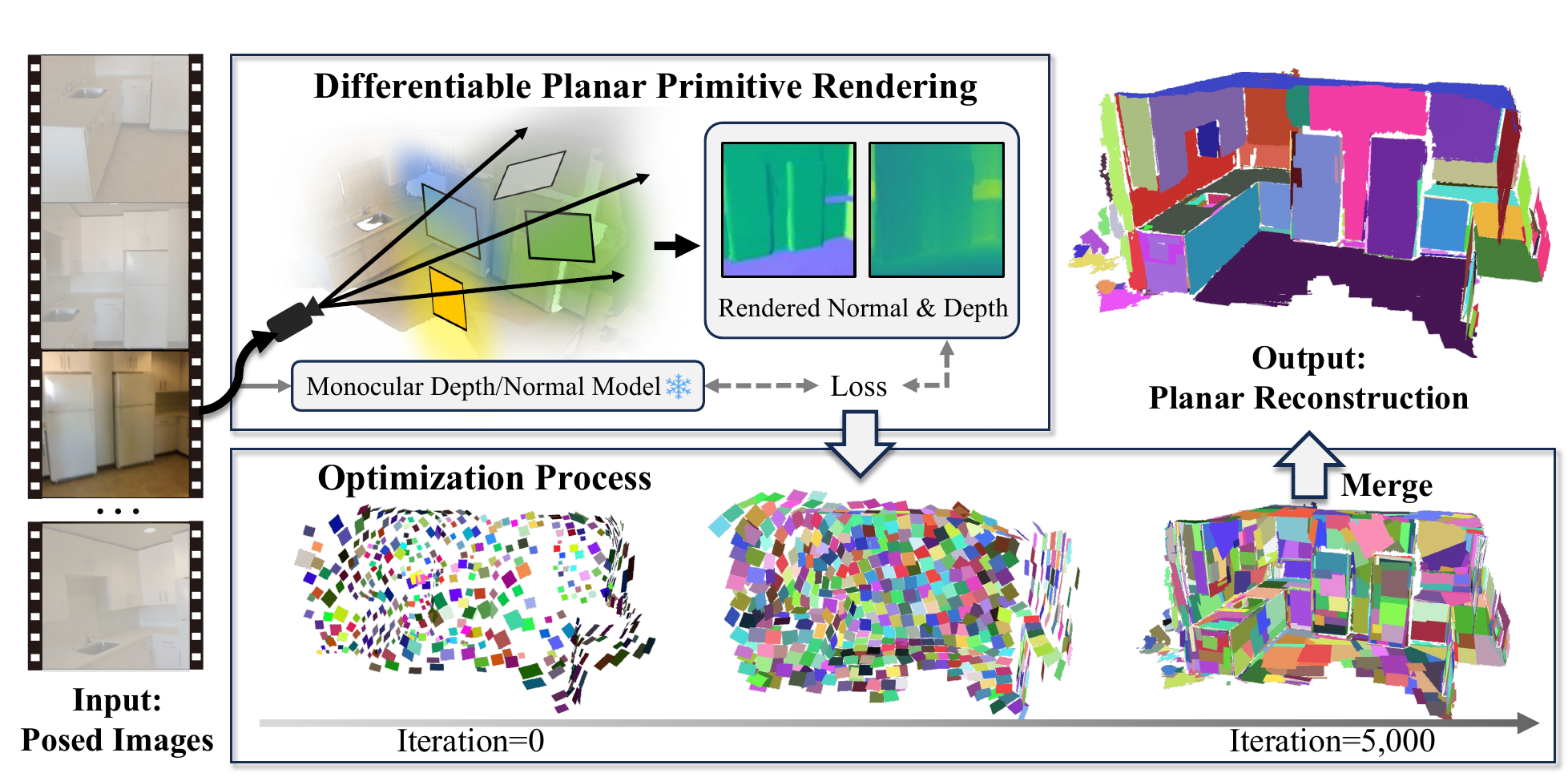}
   \caption{\textbf{Illustration of our proposed \method.} Given a set of posed multi-view images of indoor scenes, our method renders depth and normal maps from 3D plane primitives. Then, with the supervision of monocular cues, these 3D plane primitives are gradually optimized to recover the scene geometry and finally merged to get the planar reconstruction result.}
   \label{fig:pipeline} 
   \vspace{-6pt}
\end{figure*}

\section{The Proposed PlanarSplatting}\label{sec:method}
As shown in~\cref{fig:pipeline}, given a set of posed multi-view images, our \method can reconstruct the indoor planar scene from them by optimizing a set of learnable 3D planar primitives (\cref{subsec:plane_rep}). With the proposed Differentiable Planar Primitive Rendering (\cref{subsec:plane_rendering}), we can explicitly learn these planar primitives from a coarse initialization to accurately recover the scene geometry. These optimized 3D planar primitives are then merged to achieve the 3D plane instances for the final planar reconstruction.

\subsection{Learnable Planar Scene Representation}
\label{subsec:plane_rep}
Since our key idea is directly optimizing explicit 3D primitives for planar scene reconstruction, we first present the representation of our learnable 3D planar primitives and then introduce how we initialize the scene with these planar primitives. 

\noindent\textbf{Learnable Planar Primitive.} As shown in~\cref{fig:plane_rep}, we formulate a planar primitive $\pi$ as a 3D rectangle which is equipped with several learnable parameters including the plane center $\mathbf{p}_{\pi} \in \mathbb{R}^3$, the plane rotation $\mathbf{q}_{\pi} \in \mathbb{R}^4$ (in quaternion representation) and the plane radii $\mathbf{r}_{\pi}$. Specifically, to improve the optimization flexibility of plane shape, we use the design of double direction plane radii (Double Radii) for $\mathbf{r}_{\pi}$ as:
\begin{equation}
    \mathbf{r}_{\pi}=\{r_{\pi}^{x+},r_{\pi}^{x-},r_{\pi}^{y+},r_{\pi}^{y-} \} \in \mathbb{R}^4_+,
\end{equation}
where $r_{\pi}^{x+},r_{\pi}^{x-},r_{\pi}^{y+},r_{\pi}^{y-}$ are the radii defined on the positive/negative direction of the X-axis/Y-axis of the rectangle as shown in~\cref{fig:plane_rep}. For the sake of description, we further define the positive direction of the X-axis and Y-axis of the 3D planar primitive as two orthogonal unit vectors $\mathbf{v}^{x}_{\pi}, \mathbf{v}^{y}_{\pi} \in \mathbb{R}^3$ which can be calculated as:
\begin{equation}
    \mathbf{v}^{x}_{\pi} = \mathbf{R}(\mathbf{q}_{\pi}){[1,0,0]}^{\top}, \quad \mathbf{v}^{y}_{\pi} = \mathbf{R}(\mathbf{q}_{\pi}){[0,1,0]}^{\top},
\end{equation}
where $\mathbf{R}(\mathbf{q}_{\pi}) \in \mathbb{R}^{3 \times 3}$ means the rotation matrix of the quaternion $\mathbf{q}_{\pi}$. Similarly, the normal of the planar primitive $\mathbf{n}_{\pi} \in \mathbb{R}^3$ can be calculated as:
\begin{equation}
    \mathbf{n}_{\pi} = \mathbf{R}(\mathbf{q}_{\pi}){[0,0,1]}^{\top}.
\end{equation}
With these learnable parameters, the 3D planar primitive can be moved to align with the potential scene surface and deformed to fit the surface shape during optimization. 

\noindent\textbf{Scene Initialization.}
We use monocular depth from recent foundation models~\cite{Metric3Dv2} to fast initialize our 3D planar primitives at the beginning of optimization. Specifically, we use depths from Metric3Dv2~\cite{Metric3Dv2} to get a very coarse scene geometry. Then we randomly sample 2,000 points on the coarse mesh to achieve the plane centers of our 3D planar primitives. The initial radii of each primitive $\pi$ is set to $0.5Dist(\pi)$. Here, $Dist(\pi)$ means the distance closest to $\pi$ to its neighbors. We use the normal direction on the coarse mesh to initialize the plane rotation. At the bottom of~\cref{fig:pipeline}, we show an example of our initial planar primitives. With such a coarse initialization, our \method can finally reconstruct the accurate and complete scene surface.

\subsection{Differentiable Planar Primitive Rendering}
\label{subsec:plane_rendering}
To optimize the learnable planar primitives $\Pi={\{ \pi_i \} }_{i=1}^{K}$, we introduce the differentiable planar primitive rendering with a carefully designed plane splatting function which enables the planar primitives to accurately fit the scene geometry with the supervision only from 2D multi-view images.

\noindent\textbf{Ray-to-Plane Intersection.}
To project the 3D planar primitives to the 2D image space, we first calculate the intersections between planar primitives and the rays emitted from image pixels. Specifically, given a ray $\mathbf{r}=\{ \mathbf{o}, \mathbf{d} \}$ starting from the camera center $\mathbf{o} \in \mathbb{R}^3$ with direction $\mathbf{d} \in \mathbb{R}^3$, its intersection $\mathbf{x}_{\pi}^{\mathbf{r}} \in \mathbb{R}^3$ to one planar primitive $\pi$ can be calculated as:
\begin{equation}\label{eq:ray-plane-intersection}
    \mathbf{x}_{\pi}^{\mathbf{r}} = \mathbf{o} + \frac{(\mathbf{p}_{\pi} - \mathbf{o} \cdot \mathbf{n}_{\pi})}{\mathbf{d} \cdot \mathbf{n}_{\pi}} \mathbf{d},
\end{equation}
where $\mathbf{p}_{\pi}$ and $\mathbf{n}_{\pi}$ are the center and the normal of the planar primitive $\pi$. 

\begin{figure}
  \centering
   \includegraphics[width=1.0\linewidth]{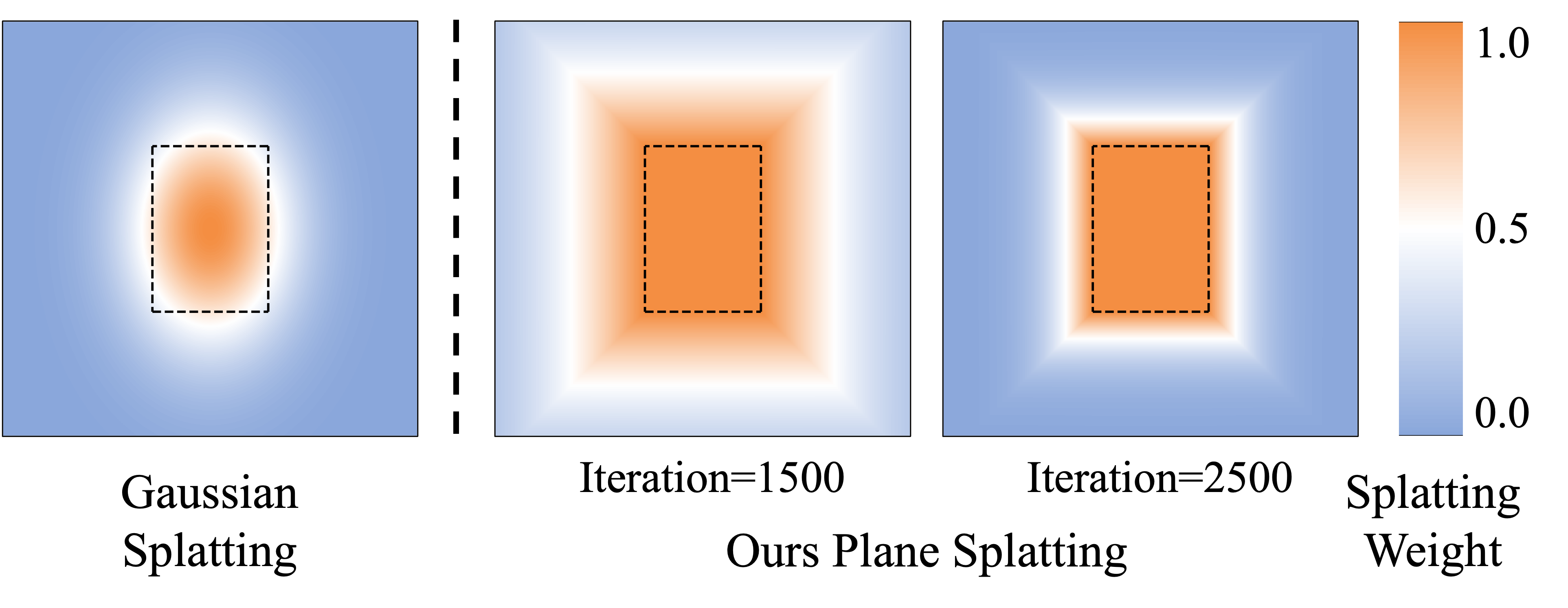}
   \caption{\textbf{Illustration of the proposed plane splatting function.} Naive Gaussian Splatting can not effectively approximate the boundary of our rectangular plane primitive (shown in black dashed border). In contrast, our proposed plane splatting function can approximate the boundary of the rectangle as the number of iterations increases, allowing our 3D planar primitives to better fit the surface of the scene.}
   \label{fig:plane_splat}
   \vspace{-6pt}
\end{figure}

\begin{figure}
  \centering
  \begin{subfigure}[c]{0.32\linewidth}
  \centering
    \includegraphics[width=1.0\linewidth]{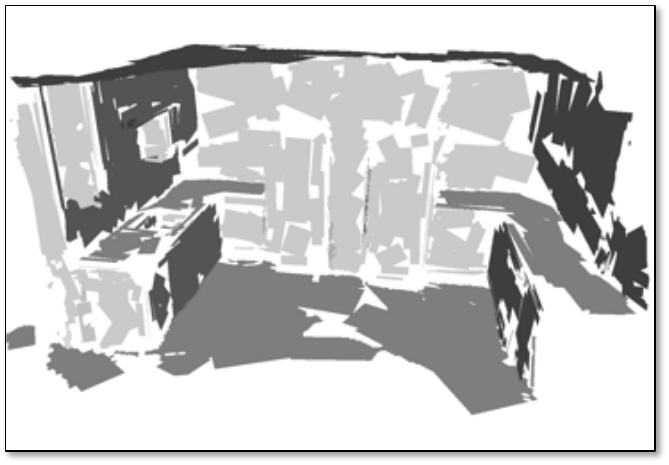}
    \caption{w/ GS Splatting}
  \end{subfigure}
  \hfill
  \begin{subfigure}[c]{0.32\linewidth}
  \centering
    \includegraphics[width=1.0\linewidth]{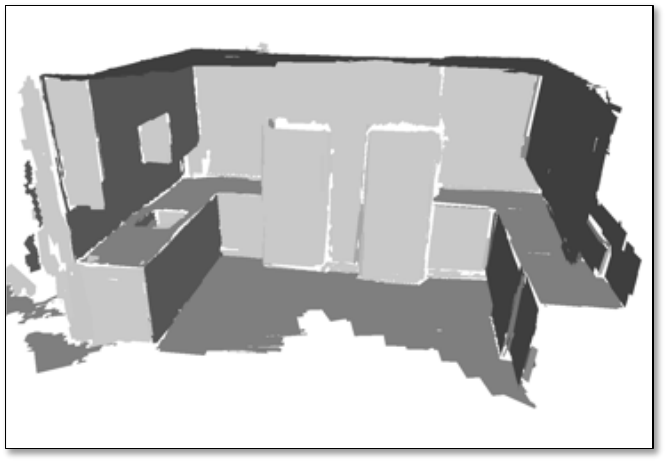}
    \caption{w/ Plane Splatting}
  \end{subfigure}
  \hfill
  \begin{subfigure}[c]{0.32\linewidth}
  \centering
    \includegraphics[width=1.0\linewidth]{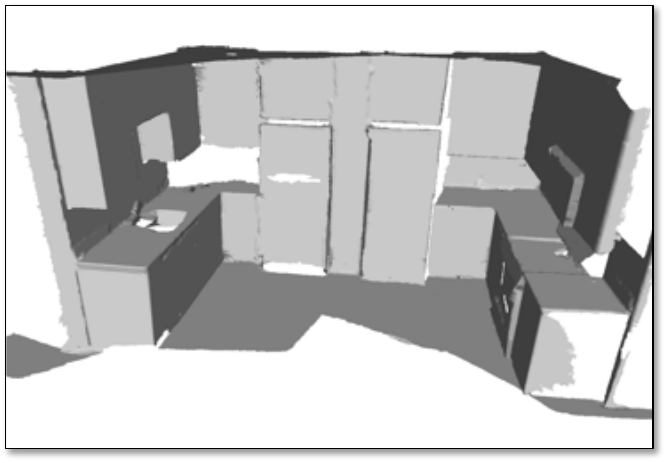}
    \caption{GT Mesh}
  \end{subfigure}
  \caption{Reconstruction comparison with different splatting functions. `w/' means `with'.}
  \label{fig:splat_res}
  \vspace{-6pt}
\end{figure}

\noindent\textbf{Plane Splatting Function.}
After achieving the ray-to-plane intersection $\mathbf{x}_{\pi}^{\mathbf{r}}$, we then calculate its splatting weight with our plane splatting function which will be used for rendering.

A vanilla selection of the splatting function is the anisotropic Gaussian function:
\begin{equation}\label{eq:gaussian-rbf}
    w(\mathbf{x}_{\pi}^{\mathbf{r}}, \pi)=\exp\left({-\frac{1}{2}(\mathbf{x}_{\pi}^{\mathbf{r}}-\mathbf{p}_{\pi})^{\top} \Sigma^{-1} (\mathbf{x}_{\pi}^{\mathbf{r}}-\mathbf{p}_{\pi})}\right),
\end{equation}
where $\mathbf{p}_{\pi}$ is the center of the planar primitive $\pi$. The covariance matrix $\Sigma$ can be calculated like~\cite{ThreeDGS-KerblKLD23,TwoDGS-HuangYC0G24}.
However, as shown in~\cref{fig:plane_splat} and~\cref{fig:splat_res}, the Gaussian-based splatting function will make ambiguous boundaries of our rectangular planar primitive leading to the degeneration of reconstruction quality.

Thus, we propose to calculate the splatting weight with a novel rectangle-based plane splatting function. To a given intersection $\mathbf{x}_{\pi}^{\mathbf{r}}$ between ray $\mathbf{r}$ and planar primitive $\pi$, we first calculate its projection distance $\mathcal{P}_{X}, \mathcal{P}_{Y} \in \mathbb{R}$ to the X-axis and Y-axis of the planar primitive as:
\begin{equation}
    \mathcal{P}_{X} = (\mathbf{x}_{\pi}^{\mathbf{r}}-\mathbf{p}_{\pi})\cdot \mathbf{v}^x_{\pi}, \quad \mathcal{P}_{Y} = (\mathbf{x}_{\pi}^{\mathbf{r}}-\mathbf{p}_{\pi})\cdot \mathbf{v}^y_{\pi}.
\end{equation}
Then, we calculate the splatting weight along the X-axis of the plane $\pi$ as:
\begin{equation}
    w_X(\mathbf{x}_{\pi}^{\mathbf{r}}) = 
    \begin{cases}
        2\sigma(5\lambda (r_{\pi}^{x+} -|\mathcal{P}_X| )), \quad \text{if}~\mathcal{P}_{X} > 0\\
        2\sigma(5\lambda (r_{\pi}^{x-} -|\mathcal{P}_X| )), \quad \text{otherwise}
    \end{cases},
\end{equation}
where $\sigma(\cdot)$ is the Sigmoid function and $\lambda$ is the hyperparameter to control the splatting weight. Similarly, we continue to calculate the splatting weight along the Y-axis of the plane $\pi$ as:
\begin{equation}
    w_Y(\mathbf{x}_{\pi}^{\mathbf{r}}) = 
    \begin{cases}
        2\sigma(5\lambda (r_{\pi}^{y+} -|\mathcal{P}_Y| )), \quad \text{if}~\mathcal{P}_{Y} > 0\\
        2\sigma(5\lambda (r_{\pi}^{y-} -|\mathcal{P}_Y| )), \quad \text{otherwise}
    \end{cases},
\end{equation}
where $r_{\pi}^{x+}, r_{\pi}^{x-}, r_{\pi}^{y+}, r_{\pi}^{y-}$ are the radii parameters of the plane $\pi$. At last, the final splatting weight can be calculated as:
\begin{equation}\label{eq:plane-rbf}
    w(\mathbf{x}_{\pi}^{\mathbf{r}}) = \begin{cases}
        w_X , \quad  \text{if}~w_X < w_Y \\
        w_Y, \quad \text{otherwise}
    \end{cases}.
\end{equation}
In~\cref{fig:plane_splat}, we show that with the increment of hyperparameter $\lambda$, our plane splatting function gradually approximates the shape of the rectangular plane primitive. In practice, we increase the value of $\lambda$ with an exponential function during optimization up to the maximum value of 300 as:
\begin{equation}
    \lambda = min(20e^{(-( 1 - 0.001 * ite))}, 300),
\end{equation}
where $ite$ means the iteration number during optimization.

\noindent\textbf{Blending Composition.} For all ray-to-plane intersections, we filter them with splatting weight lower than 0.0001 and then sort the remaining intersections according to their depth from near to far. Then, $M$ nearest intersections of each ray are selected for rendering ($M=30$ in this paper). Denote the selected intersections of a ray $\mathbf{r}$ as $P^{\mathbf{r}}=\{ \mathbf{x}_{\pi_{\tau(j)}}^{\mathbf{r}} \}_{j=1}^{M}$. Here, $\tau(j)$ indicates the index of plane among all planar primitives. At last, we render the depth and normal map of a certain image $\mathbf{I}$ as:
\begin{equation}
    \mathbf{D}_{\text{render}}^{\Pi}(\mathbf{r}) = \sum_{j=1}^{M} T_j w(\mathbf{x}_{\pi_{\tau(j)}}^{\mathbf{r}}) t_j, 
\end{equation}

\begin{equation}
    \mathbf{N}_{\text{render}}^{\Pi}(\mathbf{r}) = \sum_{j=1}^{M} T_j w(\mathbf{x}_{\pi_{\tau(j)}}^{\mathbf{r}}) \mathbf{n}_{\pi_{\tau(j)}},
\end{equation}
where
\begin{equation}
    T_j = \prod_{i=1}^{j-1}(1-w(\mathbf{x}_{\pi_{\tau(i)}}^{\mathbf{r}})).
\end{equation}
Here $t_j$ is the deoth of the intersection and $\mathbf{n}_{\pi_{\tau(j)}}$ is the normal of planar primitive $\pi_{\tau(j)}$. To supervise the rendered depth and normal map, we use the pretrained model of Metric3Dv2~\cite{Metric3Dv2} to predict the depth map ${\mathbf{D}}_{\text{pre}}$ and use Omnidata~\cite{omnidata-EftekharSMZ21} to predict the normal map ${\mathbf{N}}_{\text{pre}}$ of the image $\mathbf{I}$ to serve as pseudo labels. Finally, the render loss can be calculated as:
\begin{equation}
\begin{split}
    \mathcal{L}_{\text{render}}^{\Pi} = & \alpha_1 \sum_{\mathbf{r} \in \mathbf{I}} \| 1-\mathbf{N}_{\text{render}}^{\Pi}(\mathbf{r})^{\top}\mathbf{N}_{\text{pre}}(\mathbf{r})) \|_1 + \\ 
    & \alpha_1 \sum_{\mathbf{r} \in \mathbf{I}} \| \mathbf{N}_{\text{render}}^{\Pi}(\mathbf{r}) - \mathbf{N}_{\text{pre}}(\mathbf{r})) \|_1 + \\
    & \alpha_2 \sum_{\mathbf{r} \in \mathbf{I}} \| (\mathbf{D}_{\text{render}}^{\Pi}(\mathbf{r})) - \mathbf{D}_{\text{pre}}(\mathbf{r})\|_1, 
\end{split}
\label{eq:loss}
\end{equation}
where $\alpha_1=5.0$, $\alpha_2=1.0$, $\textbf{r}$ is the ray/pixel emitted from image $\mathbf{I}$.

\begin{table*}[!t]
    \centering
    \scriptsize
    \SetTblrInner{rowsep=2.0pt}     %
    \SetTblrInner{colsep=10.0pt}     %
    \caption{%
        Quantitative comparison of planar reconstruction results on the ScanNetV2~\cite{scannet-DaiCSHFN17} dataset. `P. Ann.' means using 2D/3D plane annotations in training stages.
    }
    \label{tab:scannetv2}
    \begin{tblr}{
        cells={halign=c,valign=m},  %
        column{1}={halign=l},       %
        cell{1}{3}={c=2}{},          %
        cell{1}{5}={c=3}{},          %
        cell{1}{8}={c=3}{},          %
        cell{1}{1}={r=2}{},          %
        cell{1}{2}={r=2}{},          %
        hline{1,2,3}={1-12}{},   %
        hline{1,8}={1.0pt},        %
        hline{2,3,5}={0.5pt},        %
        vline{2,3,5,8}={1-11}{},        %
    }
        Method  & P. Ann.&Geometry & & Segmentation & & & Planar & & \\
         & & Chamfer~$\downarrow$ & F-score~$\uparrow$ & VOI~$\downarrow$ & RI~$\uparrow$ & SC~$\uparrow$ & Fidelity~$\downarrow$ & Acc~$\downarrow$ & Chamfer~$\downarrow$ \\
         PlanarRecon~\cite{PlanarRecon-XieGYZJ22} & \checkmark &9.89&43.47 & 3.201& 0.919& 0.405& 18.86& 16.21&17.53 \\
         AirPlanes~\cite{AirPlanes-WatsonASQABFV24} & \checkmark &\underline{5.30} & 64.92& \textbf{2.268}& \textbf{0.957}& \textbf{0.568}& \underline{8.76}& \textbf{7.98}& \textbf{8.37}\\
         2DGS~\cite{TwoDGS-HuangYC0G24} + RANSAC & \ding{55} &14.15&31.33 & 4.030&0.924& 0.257& 40.02& 14.77&27.40 \\
         SR~\cite{simplerec} + RANSAC & \ding{55} &5.40& \underline{65.45}& 2.507 & 0.946& 0.515& 9.42& \underline{10.13}&9.78 \\
         Ours       & \ding{55} &\textbf{4.83}& \textbf{68.85}& \underline{2.502} &  \underline{0.948} &  \underline{0.532} & \textbf{6.64}& 11.76 & \underline{9.20} \\
    \end{tblr}
\end{table*}

\begin{table*}
    \centering
    \scriptsize
    \SetTblrInner{rowsep=2.0pt}     %
    \SetTblrInner{colsep=10.0pt}     %
    \caption{%
        Quantitative comparison of planar reconstruction results on the ScanNet++~\cite{scannetpp-YeshwanthLND23} dataset. `P. Ann.' means using 2D/3D plane annotations in training stages.
    }
    \label{tab:scannetpp}
    \begin{tblr}{
        cells={halign=c,valign=m},  %
        column{1}={halign=l},       %
        cell{1}{3}={c=2}{},          %
        cell{1}{5}={c=3}{},          %
        cell{1}{8}={c=3}{},          %
        cell{1}{1}={r=2}{},          %
        cell{1}{2}={r=2}{},          %
        hline{1,2,3}={1-12}{},   %
        hline{1,8}={1.0pt},        %
        hline{2,3,5}={0.5pt},        %
        vline{2,3,5,8}={1-11}{},        %
    }
        Method  & P. Ann.&Geometry & & Segmentation & & & Planar & & \\
         & &Chamfer~$\downarrow$ & F-score~$\uparrow$ & VOI~$\downarrow$ & RI~$\uparrow$ & SC~$\uparrow$ & Fidelity~$\downarrow$ & Acc~$\downarrow$ & Chamfer~$\downarrow$ \\
         PlanarRecon~\cite{PlanarRecon-XieGYZJ22} & \checkmark &17.85& 31.10&  3.542& 0.919& 0.367& 34.44& 19.35&26.90 \\
         AirPlanes~\cite{AirPlanes-WatsonASQABFV24} & \checkmark &13.75& 32.58& \underline{2.859}& \underline{0.941}& \underline{0.470}& \underline{28.16}&12.58 & \underline{20.37}\\
         2DGS~\cite{TwoDGS-HuangYC0G24} + RANSAC & \ding{55} &20.39& 26.46& 4.456& 0.927& 0.241& 55.90&16.88 &36.39 \\
         SR~\cite{simplerec} + RANSAC & \ding{55} &\underline{13.15}& \underline{35.93}& 3.013& 0.938& 0.442& 30.25& \textbf{11.62}&20.94 \\
         Ours       & \ding{55} &\textbf{9.33}& \textbf{47.04}& \textbf{2.772}& \textbf{0.946}& \textbf{0.523}& \textbf{17.24}& \underline{12.26} &\textbf{14.75} \\
    \end{tblr}
\end{table*}

\subsection{Optimization}
\label{subsec:plane_group}
\paragraph{Loss Function.} We optimize our \method with the Adam optimizer~\cite{adam-KingmaB14} for 5,000 iterations on each scene with the loss as shown in~\cref{eq:loss}.

\noindent\textbf{Plane Splitting.} During optimization, we introduce a splitting operation on planes according to the gradients of their radii to better fit the scene geometry. If the average radii gradients on the X-axis ($r^{x+}$ and $r^{x-}$) of the plane are greater than 0.2, we split the plane along the Y-axis. Similarly, we split the planes along the X-axis, if their radii gradients on the Y-axis ($r^{y+}$ and $r^{y-}$) are larger than 0.2. We conduct the splitting operation every 1,000 iterations.

\noindent\textbf{Plane Merge.} After optimization, we further merge the learned 3D plane primitives with normal angle error lower than $25^{\circ}$ and offset distance error lower than $0.1cm$. Here, offset means the projection distance from the scene center to the plane surface.

\noindent\textbf{CUDA Implementation.} For fast optimization of solid 3D planar primitives, we implement the forward and backward process of our Differentiable Planar Primitive Rendering with CUDA, which enables our \method to reconstruct one scene within 3 minutes. We will release the code of our CUDA Implementation for solid 3D planar primitives optimization after publication.

\section{Experiments}\label{sec:exp}
\subsection{Dataset, Metrics and Baselines}
\noindent\textbf{Datasets.} We evaluate our \method on two large indoor datasets including ScanNetV2~\cite{scannet-DaiCSHFN17} and ScanNet++~\cite{scannetpp-YeshwanthLND23} which provide posed images. On the ScanNetV2 dataset, we use the test split according to AirPlanes~\cite{AirPlanes-WatsonASQABFV24} which includes 100 scenes with ground truth 3D plane annotations provided by~\cite{planercnn-0012KGFK19}. On the ScanNet++ dataset, we randomly select 30 scenes for evaluation and extract 3D plane annotations from the ground truth meshes like~\cite{planercnn-0012KGFK19}.

\noindent\textbf{Evaluation Metrics.} According to PlanarRecon~\cite{PlanarRecon-XieGYZJ22}, 
we evaluate the geometry quality of all reconstructed 3D planes with Chamfer Distance and F-score. Following AirPlanes~\cite{AirPlanes-WatsonASQABFV24}, we also evaluate the reconstruction quality of Top-20 largest planes from the ground truth and use the metrics including Planar Fidelity, Planar Accuracy, and Planar Chamfer.
To evaluate 3D plane segmentation, we use the metrics including Variation of Information (VOI), Rand Index (RI), and Segmentation Covering (SC) like~\cite{planercnn-0012KGFK19,PlanarRecon-XieGYZJ22}. 

\noindent\textbf{Baselines.} Since our method is purely built upon geometry cues, we mainly compare our \method with those geometry-based methods including SR+RANSAC~\cite{simplerec} and 2DGS+RANSAC~\cite{TwoDGS-HuangYC0G24}. These baselines build dense scene meshes from multi-view images at first and then extract 3D planes with the RANSAC algorithm from the reconstructed meshes. We use the RANSAC implementation provided by AirPlanes~\cite{AirPlanes-WatsonASQABFV24} for all these geometry-based baselines. Besides, we also compare our \method with some plane annotation based methods that use 2D/3D plane labels/priors in their training stage including PlanarRecon~\cite{PlanarRecon-XieGYZJ22} and AirPlanes~\cite{AirPlanes-WatsonASQABFV24}, and report their results for reference.

\begin{figure*}
  \centering
  \begin{subfigure}[c]{0.24\linewidth}
  \centering
    \includegraphics[height=3.8\linewidth]{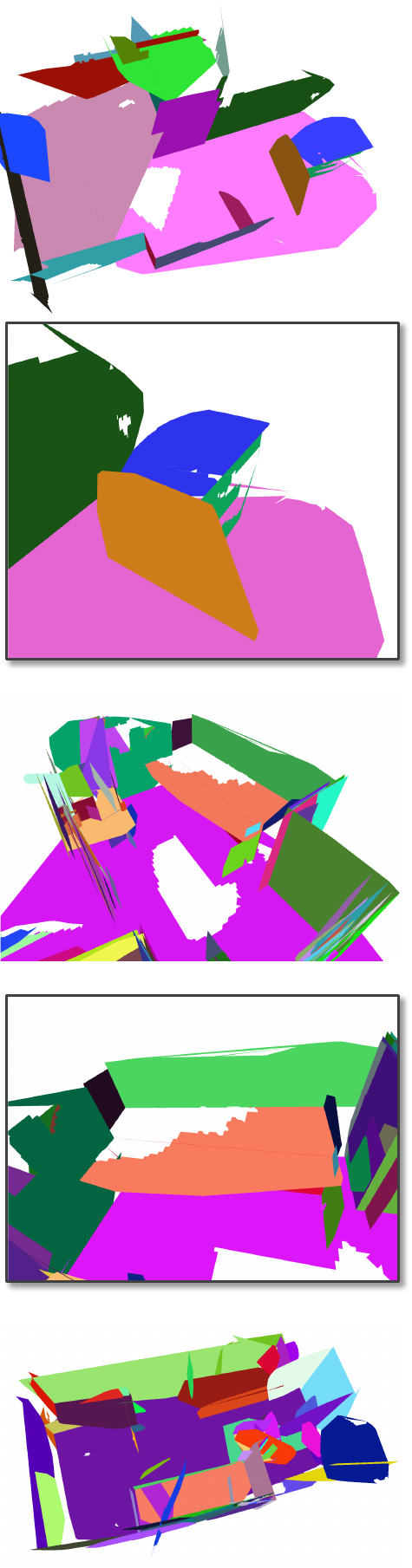}
    \caption{PlanarRecon~\cite{PlanarRecon-XieGYZJ22}}
    \label{fig:exp-compare-pr}
  \end{subfigure}
  \begin{subfigure}[c]{0.24\linewidth}
  \centering
    \includegraphics[height=3.8\linewidth]{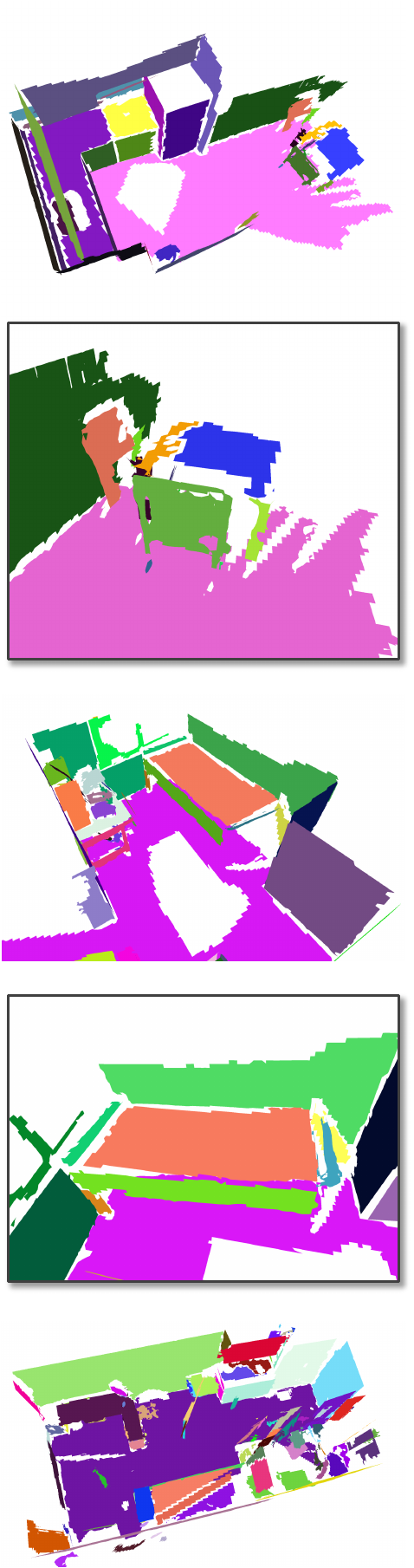}
    \caption{AirPlanes~\cite{AirPlanes-WatsonASQABFV24}}
    \label{fig:exp-compare-air}
  \end{subfigure}
  \begin{subfigure}[c]{0.24\linewidth}
  \centering
    \includegraphics[height=3.8\linewidth]{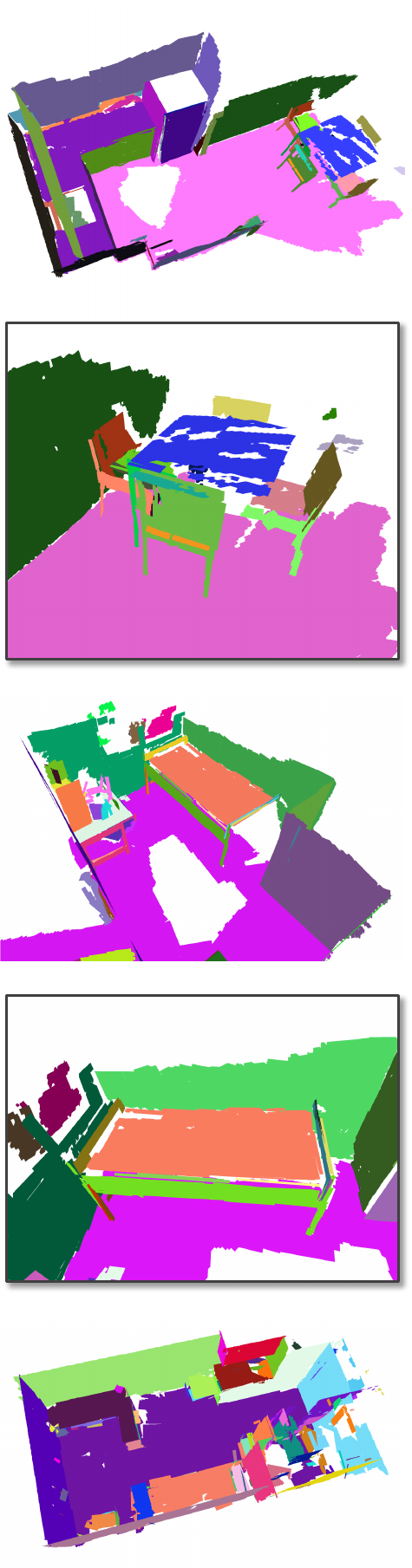}
    \caption{Ours}
    \label{fig:exp-compare-ours}
  \end{subfigure}
  \begin{subfigure}[c]{0.24\linewidth}
  \centering
    \includegraphics[height=3.8\linewidth]{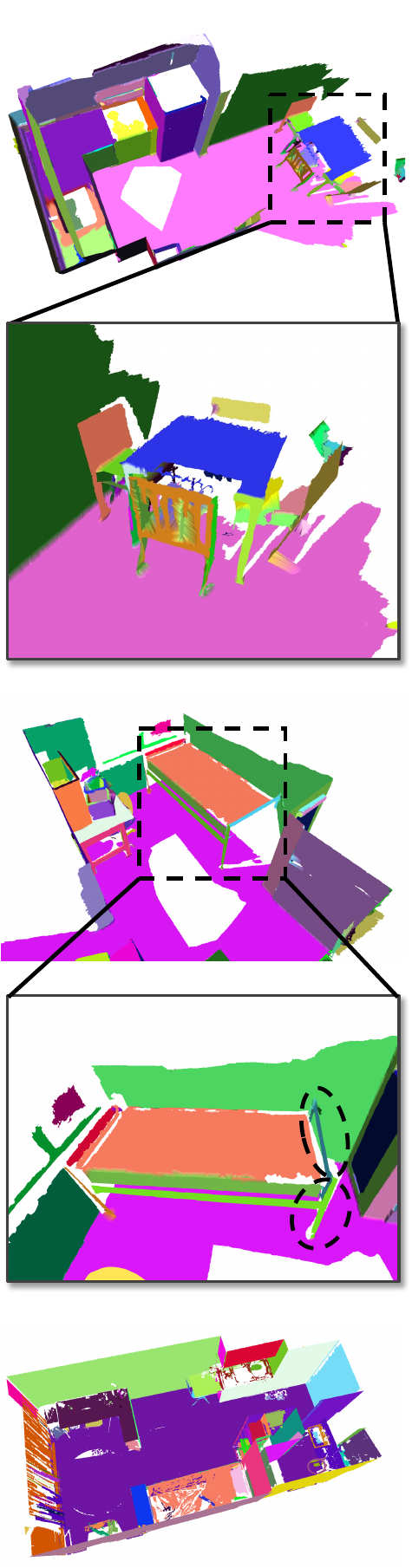}
    \caption{Ground Truth}
    \label{fig:exp-compare-gt}
  \end{subfigure}
  \caption{Qualitative comparison on the ScanNetV2 (rows 1-4) and ScanNet++ (last row) datasets.}
  \label{fig:exp-compare}
\end{figure*}

\subsection{Comparisons with Baselines}
\noindent\textbf{Quantitative Results.} We first evaluate our \method on the ScanNetV2~\cite{scannet-DaiCSHFN17} dataset. As shown in~\cref{tab:scannetv2}, our method achieves the best geometry performance in the metric of Chamfer and F-score compared to all baselines. To the results of Segmentation and Planar, our method outperforms geometry-based baselines in most metrics and is comparable to the state-of-the-art AirPlanes which uses plane embeddings trained on the same ScanNetV2 dataset. In~\cref{tab:scannetpp}, we show the results on the ScanNet++~\cite{scannetpp-YeshwanthLND23} dataset. Benefiting from our novel design for differentiable plane rendering and optimization via monocular cues provided by modern foundation models, our \method still achieves the best Geometry performance among all baselines. Furthermore, our method outperforms both geometry-based and plane annotation based methods in most metrics of Segmentation and Planar, demonstrating the robustness and superiority of our proposed \method.

\noindent\textbf{Qualitative Results.} In~\cref{fig:exp-compare}, we show the comparisons of PlanarRecon~\cite{PlanarRecon-XieGYZJ22}, AirPlanes~\cite{AirPlanes-WatsonASQABFV24} and our \method. PlanarRecon reconstructs coarse 3D planes from input multi-view images while our \method accurately reconstructs scene geometry via directly optimizing 3D plane primitives in the whole 3D space. With the help of a plane embedding model, AirPlanes achieves more semantic plane segmentation from co-plane regions. However, as shown in the zoom-in results in \cref{fig:exp-compare}, AirPlanes loses many geometry details in the scene. In contrast, our \method successfully reconstructs detailed structures such as the chairs and the legs of the bed, resulting in both accuracy and completeness in results. These comparisons effectively indicate the superiority of our method.

\subsection{Ablation Studies}
We verify the design of our \method on 10 scenes randomly selected from the ScanNetV2~\cite{scannet-DaiCSHFN17} dataset. We optimize 3D plane primitives with 5,000 iterations in default.

\noindent\textbf{Plane Initialization.} We first assess the sensitivity of the plane initialization strategy of our \method. We ablate the used Metric3D~\cite{Metric3Dv2} initialization to the sphere initialization, in which 3D plane primitives are initialized by setting their radii as 0.05 and are placed on the bounding sphere of the scene with their normals pointing to the center of the scene. As reported in the top two rows of \cref{tab:ablation}, initialization with Metric3D~\cite{Metric3Dv2} mainly improves our method to be faster converged, while sphere initialization could also obtain promising results with more iterations.

\begin{table}
    \centering
    \scriptsize
    \SetTblrInner{rowsep=2.0pt}     %
    \SetTblrInner{colsep=12.0pt}     %
    \caption{%
        Ablation studies of the proposed \method on the ScanNetV2 dataset. `w/' means `with' and `w/o' means `without'.
    }
    \label{tab:ablation}
    \begin{tblr}{
        cells={halign=c,valign=m},  %
        column{1}={halign=l},       %
        hline{1,2}={1-12}{},       %
        hline{1}={1.0pt},         %
        hline{2,4,6}={0.5pt},        %
        hline{7}={1.0pt},         %
        vline{2}={1-10}{},       %
    }   
        & Chamfer~$\downarrow$ &F-score~$\uparrow$ &VOI~$\downarrow$\\
         w/ Sphere Init.       & 8.39& 57.53& 3.073\\
         w/ Sphere Init. (30K) & \textbf{4.58}& 69.91& 2.522\\
         w/o Double Radii      & 4.73& 70.25& 2.491\\
         w/o Plane Splitting   & 4.69& 70.30& 2.477\\
         Ours (Full)           & \underline{4.66}& \textbf{71.30}& \textbf{2.473}\\
    \end{tblr}
\end{table}

\noindent\textbf{Plane Radii.} We then evaluate the design of double-direction plane radii (Double Radii) by replacing it with the vanilla single-direction plane radii. It means that we learn one radius at each axis of the plane primitive. As shown in~\cref{tab:ablation}, optimizing without Double Radii leads to a decrease in the performance of geometry and segmentation, demonstrating the effectiveness of our Double Radii.

\noindent\textbf{Plane Splitting.} As shown in the last two rows of~\cref{tab:ablation}, applying the Plane Splitting in optimization can further improve the performance in both geometry and segmentation.

\begin{table}
    \centering
    \scriptsize
    \SetTblrInner{rowsep=2.0pt}     %
    \SetTblrInner{colsep=3.3pt}     %
    \caption{%
        Quantitative comparison of novel view synthesis on the ScanNetV2~\cite{scannet-DaiCSHFN17} dataset. `Plane' means time for our plane optimization. `GS' means time for 2D/3D Gaussian optimization. `$\#$P' means the average number of Gaussian points.
    }
    \label{tab:nvs}
    \begin{tblr}{
        cells={halign=c,valign=m},  %
        column{1}={halign=l},       %
        cell{1}{1-4}={r=2}{},          %
        cell{1}{8}={r=2}{},          %
        cell{1}{5}={c=3}{},          %
        hline{1,2,3}={1-12}{},       %
        hline{1}={1.0pt},         %
        hline{3,5}={0.5pt},        %
        hline{7}={1.0pt},         %
        vline{2,5,8}={1-10}{},       %
    }   
         & PSNR~$\uparrow$ &SSIM~$\uparrow$ &LIPPS~$\downarrow$& Avg. Time (min)& & &$\#$P\\
         & & & & Plane& GS& Total&\\
         3DGS~\cite{ThreeDGS-KerblKLD23} & 24.417 & 0.781 & 0.321& -& 12.2& 12.2&1.27M\\
         Ours+3DGS~\cite{ThreeDGS-KerblKLD23} & \textbf{25.471} & \textbf{0.816} & \textbf{0.296} & 2.5 & 3.1& \textbf{5.6}&\textbf{0.37M}\\
         2DGS~\cite{TwoDGS-HuangYC0G24} & 24.766 & 0.796 & 0.323 & -& 14.2& 14.2&0.76M\\
         Ours+2DGS~\cite{TwoDGS-HuangYC0G24} & \textbf{25.380} & \textbf{0.813} & \textbf{0.296} & 2.5&4.8&\textbf{7.3}&\textbf{0.37M}\\
    \end{tblr}
\end{table}

\begin{figure}
  \centering
  \begin{subfigure}[c]{0.43\linewidth}
  \centering
    \includegraphics[width=1.0\linewidth]{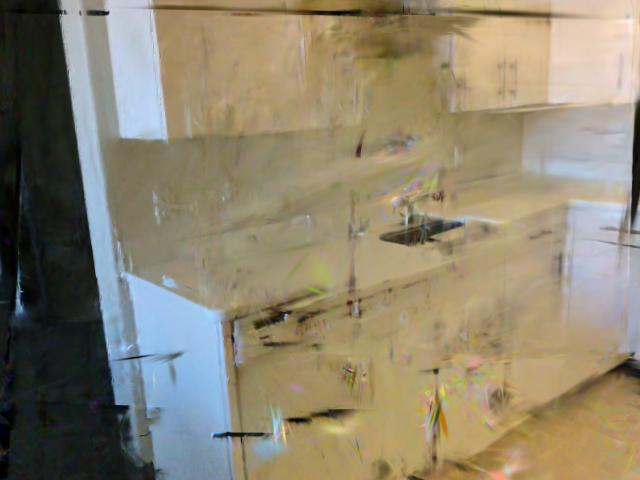}
    \caption{3DGS~\cite{ThreeDGS-KerblKLD23}}
    \label{fig:exp-nvs-3dgs}
  \end{subfigure}
  \begin{subfigure}[c]{0.43\linewidth}
  \centering
    \includegraphics[width=1.0\linewidth]{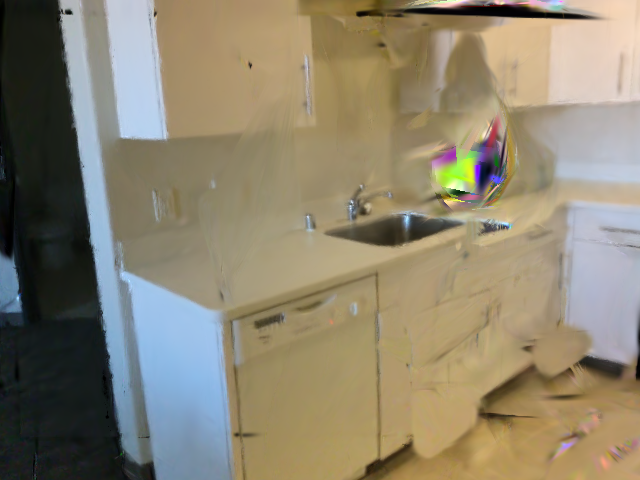}
    \caption{2DGS~\cite{TwoDGS-HuangYC0G24}}
    \label{fig:exp-nvs-2dgs}
  \end{subfigure}
  
  \begin{subfigure}[c]{0.43\linewidth}
  \centering
    \includegraphics[width=1.0\linewidth]{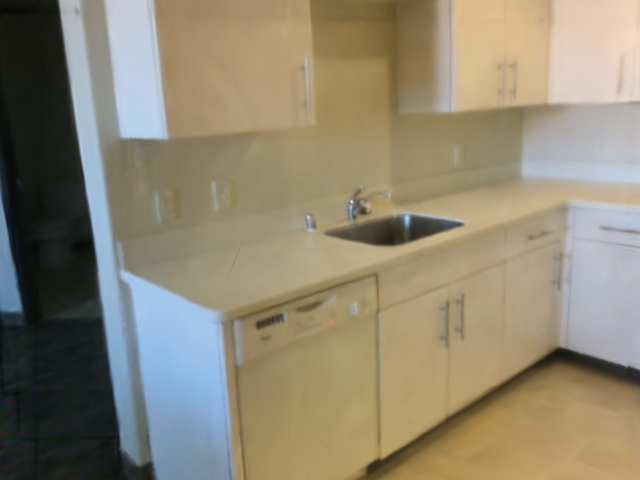}
    \caption{Ours+3DGS~\cite{ThreeDGS-KerblKLD23}}
    \label{fig:exp-nvs-ours}
  \end{subfigure}
  \begin{subfigure}[c]{0.43\linewidth}
  \centering
    \includegraphics[width=1.0\linewidth]{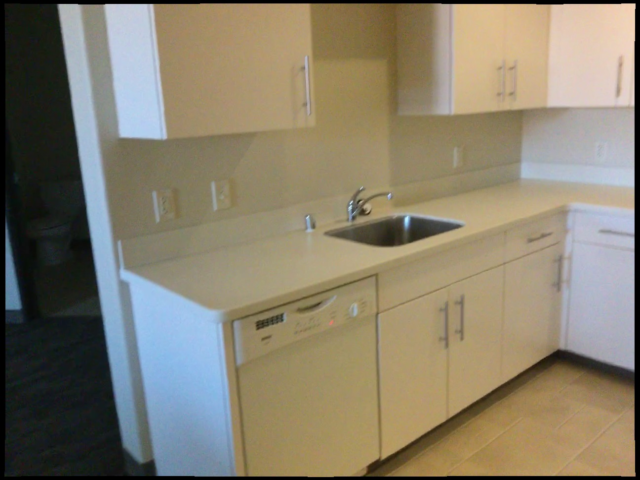}
    \caption{Ground Truth}
    \label{fig:exp-nvs-gt}
  \end{subfigure}
  \caption{Qualitative comparison of novel view synthesis on the ScanNetV2 dataset~\cite{scannet-DaiCSHFN17}. With initialization from our fast planar reconstruction, we significantly improve the rendering result of 3DGS~\cite{ThreeDGS-KerblKLD23}.}
  \label{fig:exp-nvs}
  \vspace{-6pt}
\end{figure}

\subsection{PlanarSplatting for Novel View Synthesis}
Benefiting from our accurate and fast planar reconstruction, we show that our \method can seamlessly integrate with recent Gaussian Splatting methods for more efficient and better quality novel view synthesis in indoor scenes. We select two typical methods (3DGS~\cite{ThreeDGS-KerblKLD23} and 2DGS~\cite{TwoDGS-HuangYC0G24}) for evaluation and compare them with two variants including `Ours+3DGS' and `Ours+2DGS'. Specifically, for our two variants, we directly sample points from our reconstructed 3D plane primitives to use as the initialization of these Gaussian Splatting methods. Different from the original 3DGS and 2DGS, we fix the position of 3D points and exclude the densification operation when optimizing our two variants. We conducted the experiments using the same scenes as those used in the ablation study. As shown in~\cref{tab:nvs}, our two variants `Ours+3DGS' and `Ours+2DGS' significantly outperform the original 3DGS and 2DGS in all metrics with even less optimization time and Gaussian points, demonstrating the superiority and potential of our \method. In~\cref{fig:exp-nvs}, we further show that `Ours+3DGS' effectively improves the rendering quality on the scene from the ScanNetV2 dataset.

\subsection{Limitations and Future Work}
Although our \method can reconstruct the accurate indoor planar surface, it is not suitable for complex shapes such as curved surfaces. We leave this challenging problem in our future work for more flexible geometric modeling.

\section{Conclusion}\label{sec:conclusion}
In this paper, we present \method, a novel approach for multi-view 3D reconstruction of indoor scenes. By formulating the problem through differentiable rendering with plane splatting, we demonstrate the powerful capabilities of 3D planar representation for both accurate geometry reconstruction and compact structural scene modeling. Our efficient CUDA implementation enables ultrafast 3D surface reconstruction, allowing comprehensive evaluation across over 100 scenes within hours using a single GPU. Furthermore, the seamless integration of \method with Gaussian Splatting significantly enhances both the quality and efficiency of indoor novel view synthesis, highlighting the broader potential of our approach and the inherent advantages of 3D planar representations for indoor scene understanding.

{
\small
\bibliographystyle{ieeenat_fullname}
\bibliography{ref.bib}
}

\clearpage
\appendix
\renewcommand\thesection{\Alph{section}}
\renewcommand\thefigure{S\arabic{figure}}
\renewcommand\thetable{S\arabic{table}}
\renewcommand\theequation{S\arabic{equation}}
\setcounter{figure}{0}
\setcounter{table}{0}
\setcounter{equation}{0}
\setcounter{page}{1}
\maketitlesupplementary
\section*{Appendix}

\section{More Details of \method}\label{appendix:overview}
\subsection{Data Preparation for Optimization}
On both the ScanNetV2~\cite{scannet-DaiCSHFN17} and ScanNet++~\cite{scannetpp-YeshwanthLND23} datasets, we used images sized at $480 \times 640$ for our \method. On the ScanNetV2 dataset, we sample images for optimization from the original video at intervals of 8 frames. On the ScanNet++ dataset, we sample images for optimization from the original video at intervals of 10 frames. 

\subsection{Optimization Details}
\begin{figure}[!h]
     \centering
     \begin{subfigure}[b]{0.46\textwidth}
         \centering
         \includegraphics[width=\textwidth]{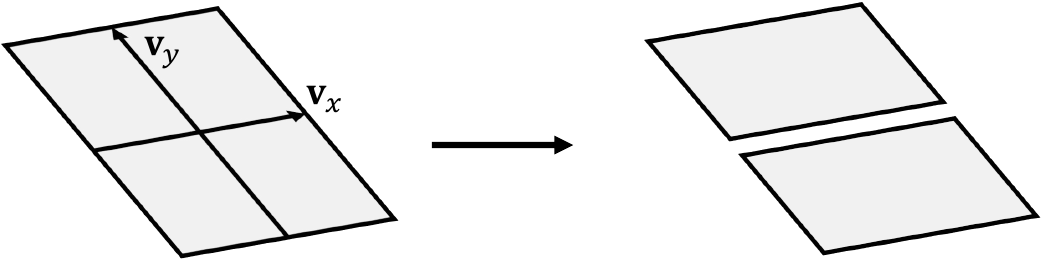}
         \caption{Split Plane along X-axis ($v_x$).}
     \end{subfigure}
     \hfill
     \begin{subfigure}[b]{0.46\textwidth}
         \centering
         \includegraphics[width=\textwidth]{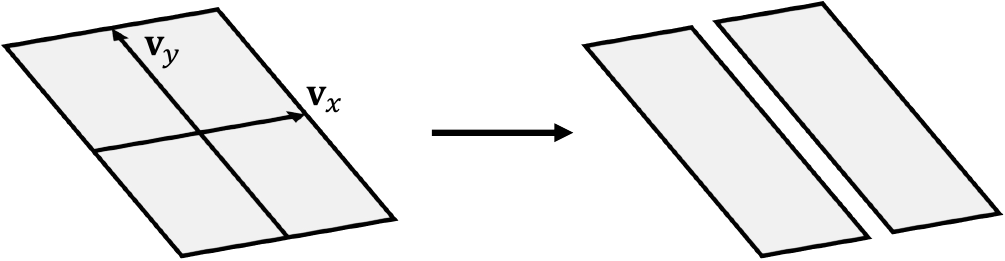}
         \caption{Split Plane along Y-axis ($v_y$).}
     \end{subfigure}
     \hfill
    \caption{Illustration of Plane Splitting.}
    \label{fig:supp_split_example}
\end{figure}

The learning rates of the learnable plane centers, plane radii, and plane rotation are all fixed at 0.001. We introduce Plane Splitting during optimization to better fit the scene geometry. In~\cref{fig:supp_split_example}, we present two examples to explain our Plane Splitting operation along the X-axis and Y-axis of the 3D plane primitive.

\section{More Qualitative Results}
\subsection{Novel View Synthesis}
\begin{figure}
  \centering
   \includegraphics[width=1.0\linewidth]{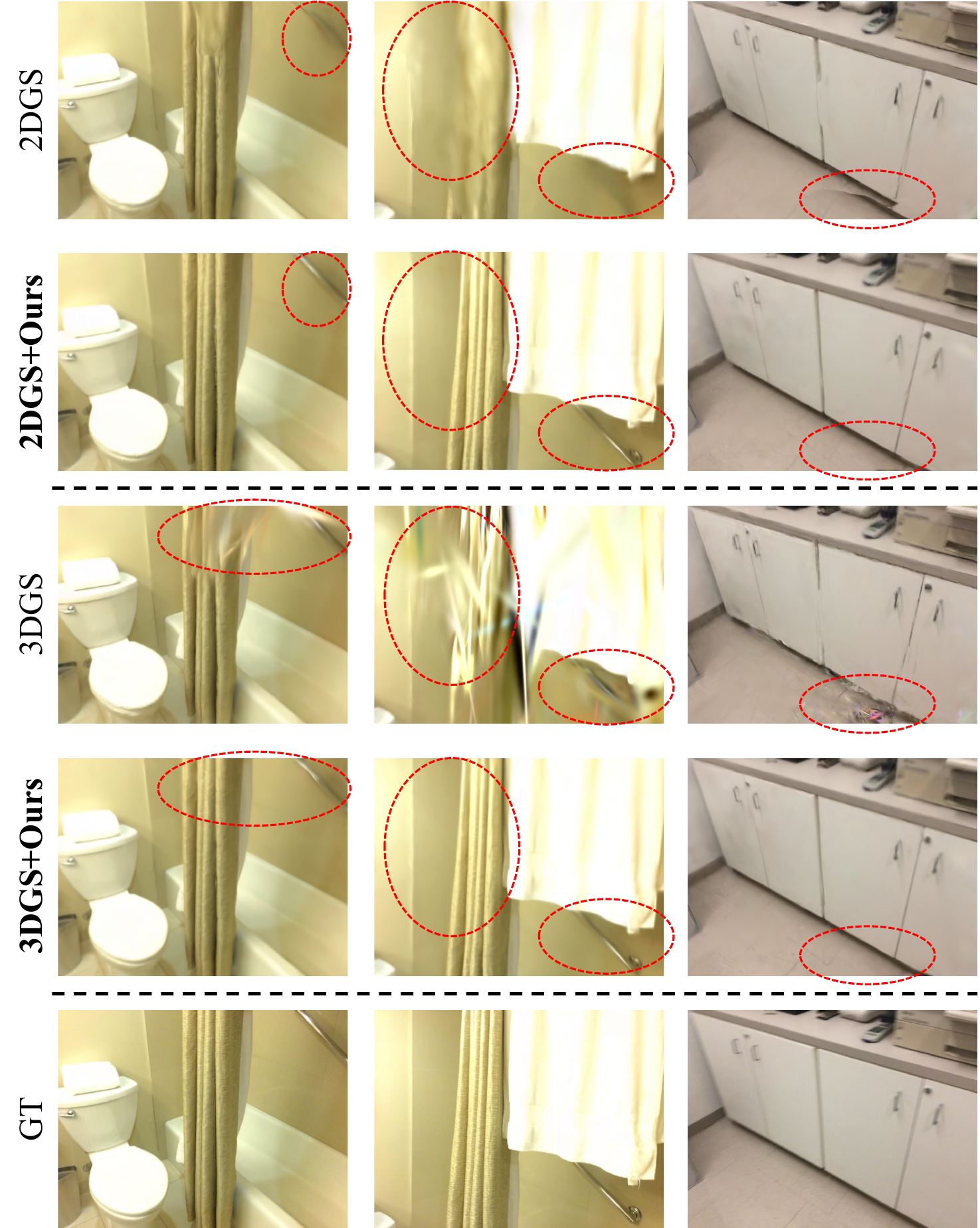}
   \caption{More qualitative comparison of novel view synthesis on the ScanNetV2~\cite{scannet-DaiCSHFN17} dataset.}
   \label{fig:supp-nvs} 
\end{figure}

In~\cref{fig:supp-nvs}, we show more novel view synthesis results on the ScanNetV2~\cite{scannet-DaiCSHFN17} dataset. By combining our \method with 2DGS~\cite{TwoDGS-HuangYC0G24} and 3DGS~\cite{ThreeDGS-KerblKLD23}, the rendering results are significantly improved.

\subsection{Planar Reconstruction}
\begin{figure*}
  \centering
   \includegraphics[width=0.83\linewidth]{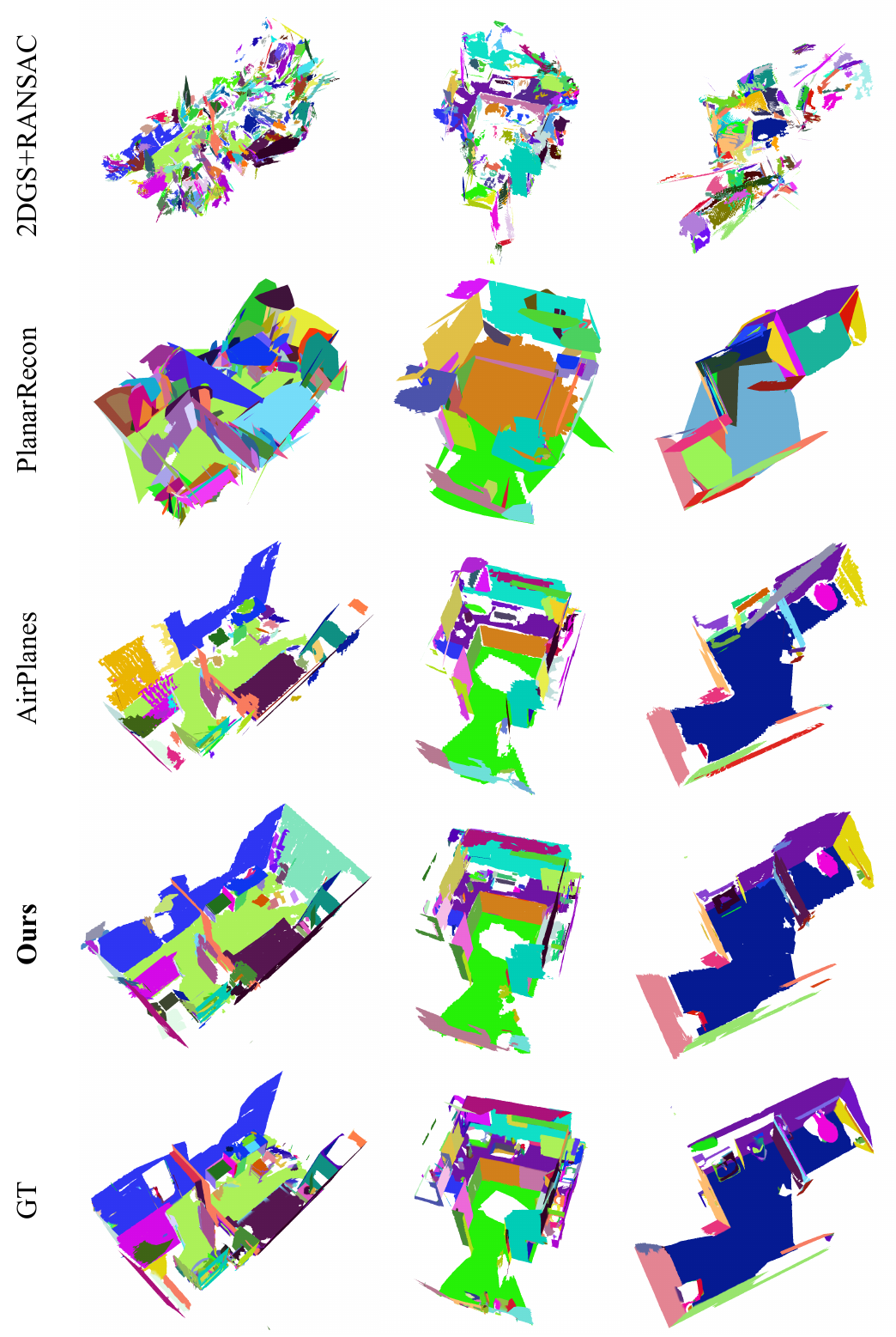}
   \caption{More qualitative comparison of planar reconstruction on the ScanNetV2~\cite{scannet-DaiCSHFN17} dataset.}
   \label{fig:supp-rec-1} 
\end{figure*}

\begin{figure*}
  \centering
   \includegraphics[width=0.85\linewidth]{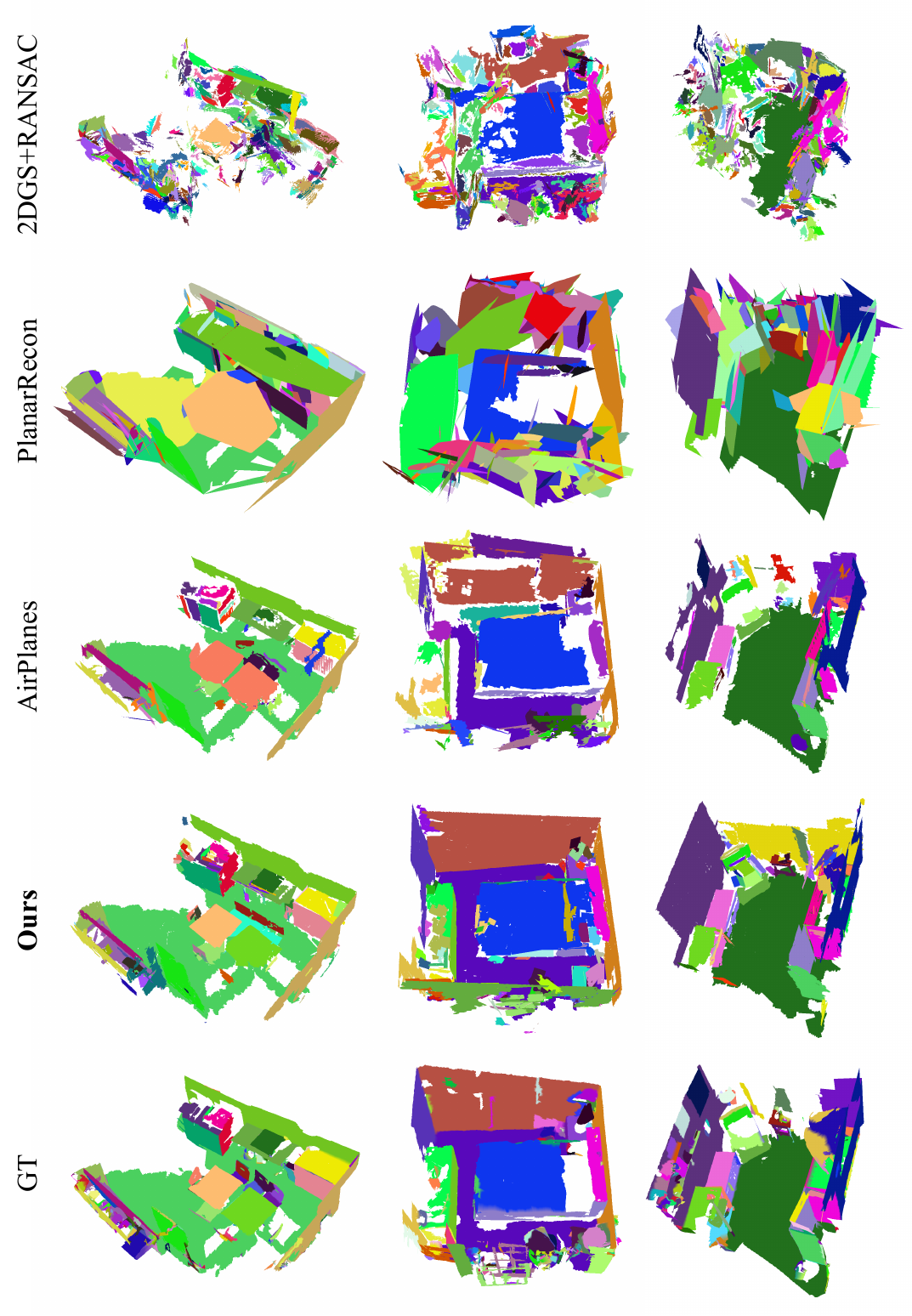}
   \caption{More qualitative comparison of planar reconstruction on the ScanNetV2~\cite{scannet-DaiCSHFN17} dataset.}
   \label{fig:supp-rec-2} 
\end{figure*}

\begin{figure*}
  \centering
   \includegraphics[width=0.88\linewidth]{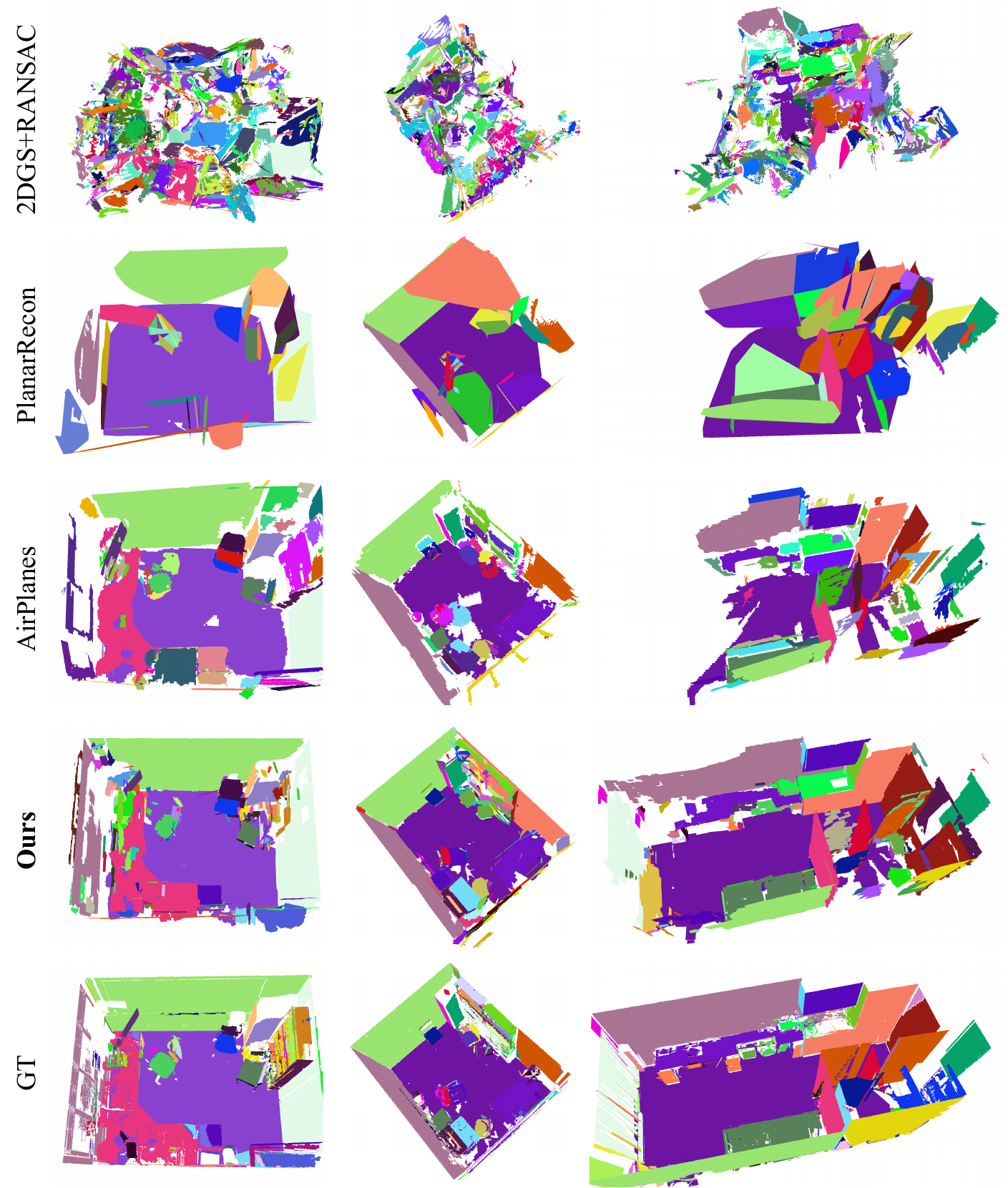}
   \caption{More qualitative comparison of planar reconstruction on the ScanNet++~\cite{scannetpp-YeshwanthLND23} dataset.}
   \label{fig:supp-rec-3} 
\end{figure*}

In~\cref{fig:supp-rec-1},~\cref{fig:supp-rec-2}, and~\cref{fig:supp-rec-3}, we show more planar reconstruction results on the ScanNetV2~\cite{scannet-DaiCSHFN17} and ScanNet++~\cite{scannetpp-YeshwanthLND23} datasets. Compared to the baselines including 2DGS~\cite{TwoDGS-HuangYC0G24}+RANSAC, PlanarRecon~\cite{PlanarRecon-XieGYZJ22} and AirPlanes~\cite{AirPlanes-WatsonASQABFV24}, our \method can achieve more accurate and complete plane reconstruction results, demonstrating the superiority of our proposed \method.

\end{document}